\documentclass[conference,a4paper]{APSIPA2025}
\usepackage{amsmath}
\usepackage{graphicx}
\usepackage{multirow}
\usepackage{threeparttable}
\usepackage{booktabs}
\usepackage{subcaption}

\usepackage{geometry}
\usepackage{fancyhdr}

\fancypagestyle{firststyle}{
  \fancyhf{}
  \fancyhead[C]{}
}

\begin{document}

%%%%%%%%%%%%%%%%%%%%%%%%%%%%%%
\title{Cross-patient Seizure Onset Zone Classification by Patient-Dependent Weight}

\author
{
\authorblockN{
Xuyang Zhao\authorrefmark{1}\authorrefmark{2}\authorrefmark{3}\authorrefmark{4}, 
Hidenori Sugano\authorrefmark{5}, 
and
Toshihisa Tanaka\authorrefmark{4}}

\authorblockA{\authorrefmark{1}
Medical Science Data-driven Mathematics Team,\\
RIKEN Center for Interdisciplinary Theoretical and Mathematical Sciences, Japan}

\authorblockA{\authorrefmark{2}
Medical Data Mathematical Reasoning Special Team,\\
RIKEN Center for Integrative Medical Sciences, Japan}

\authorblockA{\authorrefmark{3}
Department of Artificial Intelligence Medicine, Chiba University, Japan}

\authorblockA{\authorrefmark{4}
Department of Electrical Engineering and Computer Science,\\
Tokyo University of Agriculture and Technology, Japan}

\authorblockA{\authorrefmark{5}
Faculty of Medicine, Juntendo University, Japan}
}

\maketitle
\thispagestyle{firststyle}
\pagestyle{fancy}

\begin{abstract}
Identifying the seizure onset zone (SOZ) in patients with focal epilepsy is essential for surgical treatment and remains challenging due to its dependence on visual judgment by clinical experts. The development of machine learning can assist in diagnosis and has made promising progress. However, unlike data in other fields, medical data is usually collected from individual patients, and each patient has different illnesses, physical conditions, and medical histories, which leads to differences in the distribution of each patient's data. This makes it difficult for a machine learning model to achieve consistently reliable performance in every new patient dataset, which we refer to as the ``cross-patient problem.'' In this paper, we propose a method to fine-tune a pretrained model using patient-specific weights for every new test patient to improve diagnostic performance.
First, the supervised learning method is used to train a machine learning model. Next, using the intermediate features of the trained model obtained through the test patient data, the similarity between the test patient data and each training patient's data is defined to determine the weight of each training patient to be used in the following fine-tuning.
% (high similarity means high patient weight).
Finally, we fine-tune all parameters in the pretrained model with training data and patient weights. In the experiment, the leave-one-patient-out method is used to evaluate the proposed method, and the results show improved classification accuracy for every test patient, with an average improvement of more than 10\%.
\end{abstract}
%%%%%%%%%%%%%%%%%%%%%%%%%%%%%%

%%%%%%%%%%%%%%%%%%%%%%%%%%%%%%
\section{Introduction}
Epilepsy is a common neurological disease that affects 50 million people worldwide, according to the World Health Organization~\cite{world2019epilepsy}. Patients usually experience irregular epileptic seizures, which are caused by abnormal discharges of neurons and cause some problems for patients, such as spasms, rigidity, clonus, and others in their lives~\cite{fisher2014ilae, ngugi2011incidence, yaffe2015physiology}. For infant patients, epilepsy also has a negative impact on the future neurological development of the patient, making future life more difficult~\cite{korenke2004severe}. 
In current clinical diagnosis and treatment, patients first undergo a variety of medical examinations to determine their condition, among which long-term electroencephalogram (EEG) monitoring is an important basis for diagnosis~\cite{gibbs1968electro, smith2005eeg}.
During the treatment phase, medication is usually administered first to control the condition. After drug treatment, some patients can be completely seizure-free, and some patients need lifelong medication, but still some patients develop drug resistance and cannot be controlled with drugs~\cite{tomson2004medical, sinha2017predicting}. For patients whose disease cannot be controlled by medication, surgery is an optional treatment method. The seizure onset zone (SOZ) channel must be accurately located prior to surgery, and resection of the affected area in the brain may benefit patients~\cite{jobst2015resective}.
Currently, SOZ localization is performed by clinical experts using the visual judgment of intracranial electroencephalography (iEEG).
% The iEEG data is recorded by two types of electrodes of subdural grids that are placed on the cortical surface and depth electrodes that are placed in the deep cortex. This is determined by the clinical expert based on the patient's condition~\cite{gonzalez2007long}.
% After iEEG collection, the iEEG is visually judged by clinical experts. 
However, this is a difficult, time-consuming, and subjective task. Clinical experts also need long-term training to perform this diagnostic task. In order for diagnosis to be timely and accurate, high-accuracy diagnostic aids are required to identify SOZ channels.

Some methods have been proposed to assist the SOZ localization task. The proposed methods can be classified mainly into two categories: feature extraction with a classifier and end-to-end machine learning model~\cite{siddiqui2020review, hossain2019applying, islam2023epileptic}.
In the first category, time-frequency, entropy, empirical mode decomposition, etc., are used as features, and the support vector machine, random forest, Bayesian, etc., are used as classifiers.
Although these methods have achieved good performance according to the experimental results, it is difficult for them to demonstrate universality because the manually proposed feature extraction depends on the dataset.
With the development of machine learning, more and more end-to-end models have been proposed for the classification task of SOZ and non-SOZ~\cite{jha2024predictors, besheli2024real, balaji2024seizure, zhao2022seizure}.
In machine learning methods, the model is first trained by using clinically collected data. The trained model is then used to make inferences on the data from the test patient. Usually, the convolutional neural network (CNN), the recurrent neural network, and long-term memory are used as models. Learning from the training data, the model can classify the test data well, but there is an assumption here that the distribution of the training data and the test data is consistent.
Unlike data in other fields, medical data is composed of independent patients, and each patient's physical condition, illness, medical history, etc., are different, resulting in unknown differences in the patient data. Due to the distribution differences between the patients, the classification performance of the model is often poor for patients with distribution differences from the training data, which we refer to as the ``cross-patient problem.''

The cross-patient problem in the medical data makes the performance of the model inference for each patient different from another. To address the cross-patient problem in the epilepsy data, some methods have been proposed~\cite{zhang2024cross, shafiezadeh2024systematic, zhang2024efficient, wang2022cross, matsubayashi2023identification}. In these methods, some methods design a complex model to structure the learning of individual and common features of patients, or use feature-based transfer learning to adjust the data distribution. From the experimental results, the proposed method improves the cross-patient problem to a certain extent. However, both complex models and feature-based transfer learning mean more complex calculations and more parameter adjustments, which makes the methods lack generality and are not conducive to clinical application.

In this paper, we propose a method based on patient weight fine-tuning to improve the cross-patient problem existing in the epilepsy data. The method consists of two steps. The first step is to train the model using a training dataset by common supervised learning, and a one-dimensional convolutional neural model (1D-CNN) is used as the classifier for SOZ and non-SOZ classification tasks. In the second step, we fine-tune the trained model in the first step using the training data, and each patient in the training data is assigned a patient weight.
The patient weights are calculated using the output of the model's intermediate layers (last convolutional layer) between test patient data and each training patient's data. In the experiment, we calculated the similarity between the test patient and each training patient by using the maximum mean discrepancy (MMD)~\cite{gretton2012kernel} method. 
%The similarity is then processed through softmax and used as the patient weight for model fine-tuning.
The similarity is used as the patient's weight for model fine-tuning.
In general, if the similarity between the test patient and one of the training patient data is high, the weight of these patient data will increase; otherwise, it will decrease.
In the experiment, leave-one-patient-out is used to evaluate the performance of the proposed method. From the experimental results, the performance of each patient has improved under the two methods of calculating similarity (MMD and Multiscale, RBF) and the average increase of more than 10\% in accuracy.

The rest of the paper is organized as follows:
Section \uppercase\expandafter{\romannumeral2} describes the
iEEG dataset, patient weight method, and 1D-CNN model.
Section \uppercase\expandafter{\romannumeral3} describes the experimental process and results.
Section \uppercase\expandafter{\romannumeral4} describes the discussion of the proposed method, and the conclusion is presented in Section \uppercase\expandafter{\romannumeral5}.

%%%%%%%%%%%%%%%%%%%%%%%%%%%%%%
\section{Materials and methods}

\subsection{Dataset}
\begin{figure}[t]
\centerline{\includegraphics[width=1.0\columnwidth]{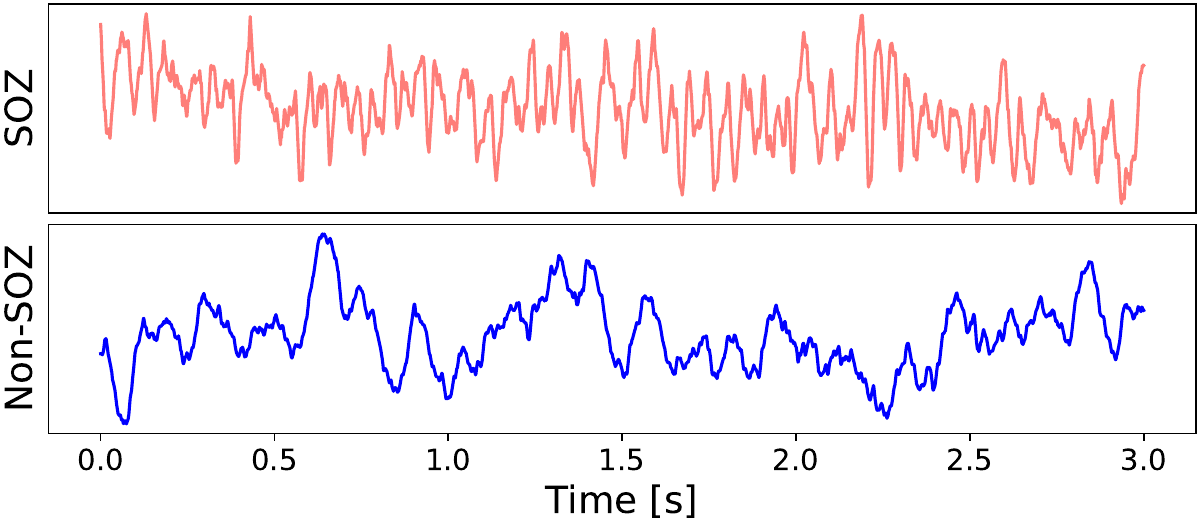}}
\caption{SOZ and non-SOZ samples in the Juntendo iEEG dataset.}
\label{fig_eeg}
\end{figure}

The dataset used to evaluate the proposed method was collected from the Epilepsy Center of Juntendo University Hospital, Tokyo, Japan. The research is approved by the Ethics Committee of the Juntendo University Hospital and the Ethics Committee of the Tokyo University of Agriculture and Technology. 
%The research is approved by the Ethics Committee.
A total of 11 patients (referred to as P01 to P11 from now on) are included in the dataset with medically intractable epilepsy caused by focal cortical dysplasia (FCD); the information for each patient is shown in Table~\ref{tab_dataset}.
In clinical collection, subdural electrodes (UNIQUE MEDICAL Co, Tokyo, Japan) 
%In clinical collection, the subdural electrodes 
are implanted to cover almost the entire surface over the FCD and the adjacent cortex.
The iEEG data collection uses a sampling frequency of 2,000 for P01-P08 and 1,000 for P09-P11, and the data with a sampling rate of 2,000 will be downsampled to 1,000.
The continuously recorded iEEG data is divided into three seconds samples, and the SOZ and non-SOZ examples are shown in Figure~\ref{fig_eeg}. Each patient contained 4,640 SOZ samples and 4,640 non-SOZ samples.

%%%%%%%%%%%%%%%%%%%%%%%%%
\begin{table*}[htbp]
\caption{Summary of information of each patient.}
\label{tab_dataset}
\begin{center}
\begin{tabular}{c | c | c | c | c | c | c | c}
\toprule
        & Age & Gender & Lesion Site & Pathology & Engel & Follow Up & Sampling Rate\\ \midrule
P01 &  5  & F  & Lt dorsal superior temporal gyrus & Cortical surface                    & Type 2B & 3 year       & 2,000 \\ \midrule
P02 & 39 & F  & Lt dorsal superior frontal gyrus     & Bottom of sulcus                  & Type 2B & 3 year       & 2,000 \\ \midrule
P03 &  5  & M & Lt cingulate gyrus                          & Bottom of sulcus                 & Type 2B & 3.5 year    & 2,000 \\ \midrule
P04 &  6  & M & Rt dorsal middle frontal gyrus       & Cortical surface                   & Type 2B & 3.5 year    & 2,000 \\ \midrule
P05 & 20 & M & Rt middle frontal gyrus                  & Cortical surface                   & Type 2A & 4.5 year    & 2,000 \\ \midrule
P06 & 15 & M & Lt superior parietal lobule              & Cortical surface                   & Type 2B & 5 year      & 2,000 \\ \midrule
P07 & 32 & M & Lt superior parietal lobule              & Bottom of sulcus                 & Type 2B & 5 year      & 2,000 \\ \midrule
P08 & 25 & M & Lt angular gyrus                             & Bottom of sulcus                 & Type 2A & 5 year      & 2,000 \\ \midrule
P09 & 38 & F  & Rt supramarginal gyrus                 & Surface and vertical cortex  & Type 2B & 5.5 year   & 1,000 \\ \midrule
P10 & 14 & F  & Rt inferior frontal gyrus                 & Cortical surface                    & Type 2B & 5.5 year   & 1,000 \\ \midrule
P11 & 13 & M & Lt angular gyrus                            & Surface and vertical cortex  & Type 2B & 5 year      & 1,000 \\
\bottomrule
\end{tabular}
\end{center}
\end{table*}

\subsection{Patient data distribution}
\begin{figure}[htbp]
    \centering
    \begin{subfigure}[b]{1.0\columnwidth}
        \includegraphics[width=1.0\columnwidth]{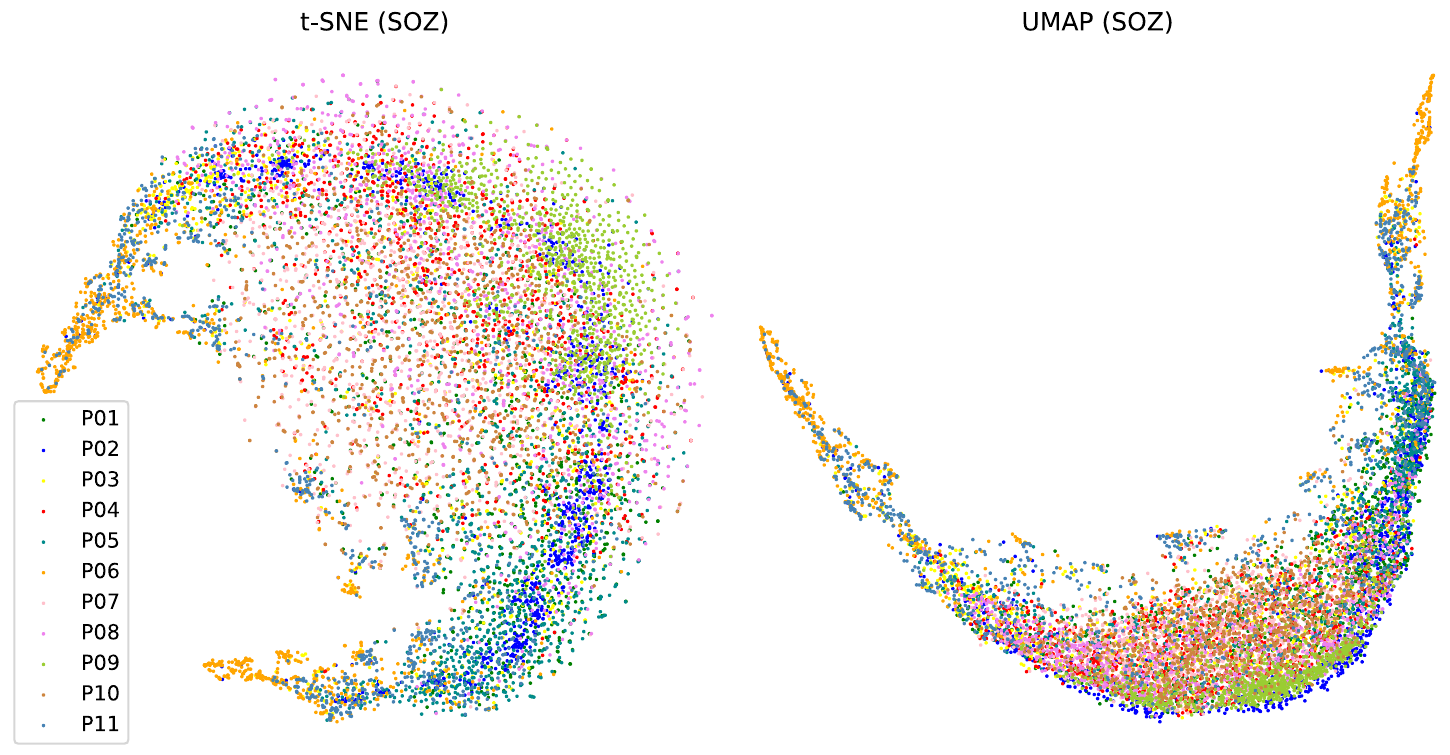}
        \caption{SOZ data distribution visualization (t-SNE \& UMAP).}
        \label{fig_soz_visualization}
    \end{subfigure}
    \hfill
    \begin{subfigure}[b]{1.0\columnwidth}
        \includegraphics[width=1.0\columnwidth]{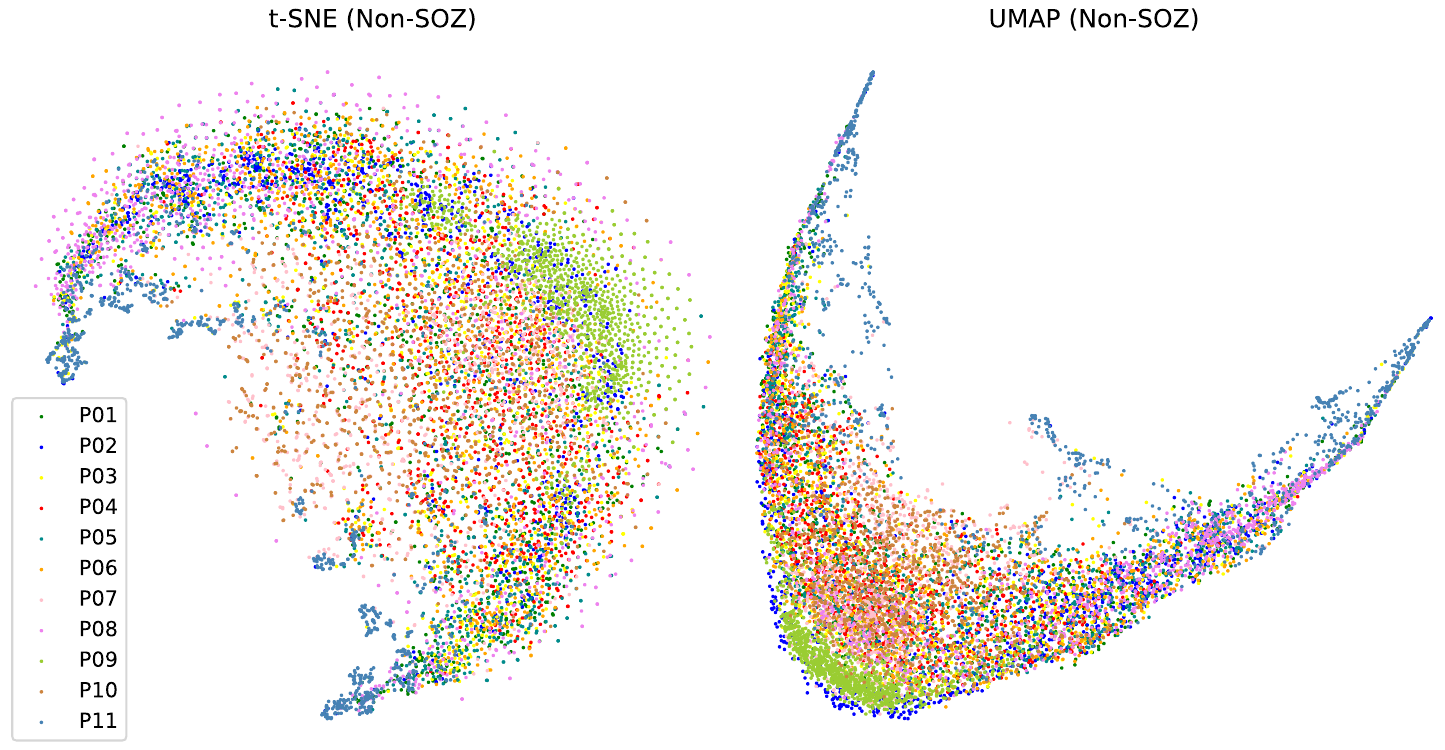}
        \caption{Non-SOZ data distribution visualization (t-SNE \& UMAP).}
        \label{fig_non_soz_visualization}
    \end{subfigure}
    \caption{Data distribution visualization.}
    \label{fig_visualization}
\end{figure}

Unlike data in other fields, medical data is composed of independent patients, and the data distribution of each patient is different due to factors such as disease condition, age, physical condition, medical history, etc.
The Juntendo dataset contains data from eleven patients in total, each of which contains 4,640 SOZ samples and 4,640 non-SOZ samples.
In order to see the distribution differences of different patient data, we use t-SNE~\cite{van2008visualizing} and UMAP~\cite{mcinnes2018umap} to visualize the two types of iEEG data separately, the results are shown in the Figure~\ref{fig_soz_visualization} and Figure~\ref{fig_non_soz_visualization}. As shown in the figures, there are certain differences in the distribution of data of different patients, such data makes it difficult to ensure that the model performs well in every test patient.

%\begin{figure}[htpb]
%\centerline{\includegraphics[width=1.0\columnwidth]{./fig/map_f.pdf}}
%\caption{SOZ data distribution visualization (t-SNE \& UMAP).}
%\label{fig_map_soz}
%\end{figure}

%\begin{figure}[htpb]
%\centerline{\includegraphics[width=1.0\columnwidth]{./fig/map_n.pdf}}
%\caption{Non-SOZ data distribution visualization (t-SNE \& UMAP).}
%\label{fig_map_non-soz}
%\end{figure}

\subsection{Convolutional neural network (CNN) models}
CNNs have demonstrated remarkable effectiveness in data classification tasks. In this work, we employ a CNN model to classify SOZ and non-SOZ data. The iEEG data are segmented into three-second samples, the model is implemented as 1D-CNN.  The SOZ data include some features, such as spikes, sharp waves, and slow waves, which differentiate them from non-SOZ data. The 1D-CNN model is particularly suited for extracting the distinctive features that separate these two data classes. The 1D-CNN model comprises ten convolutional layers followed by three fully connected layers, designed to effectively classify SOZ and non-SOZ data based on the extracted features.

\begin{figure}[htpb]
\centerline{\includegraphics[width=0.7\columnwidth]{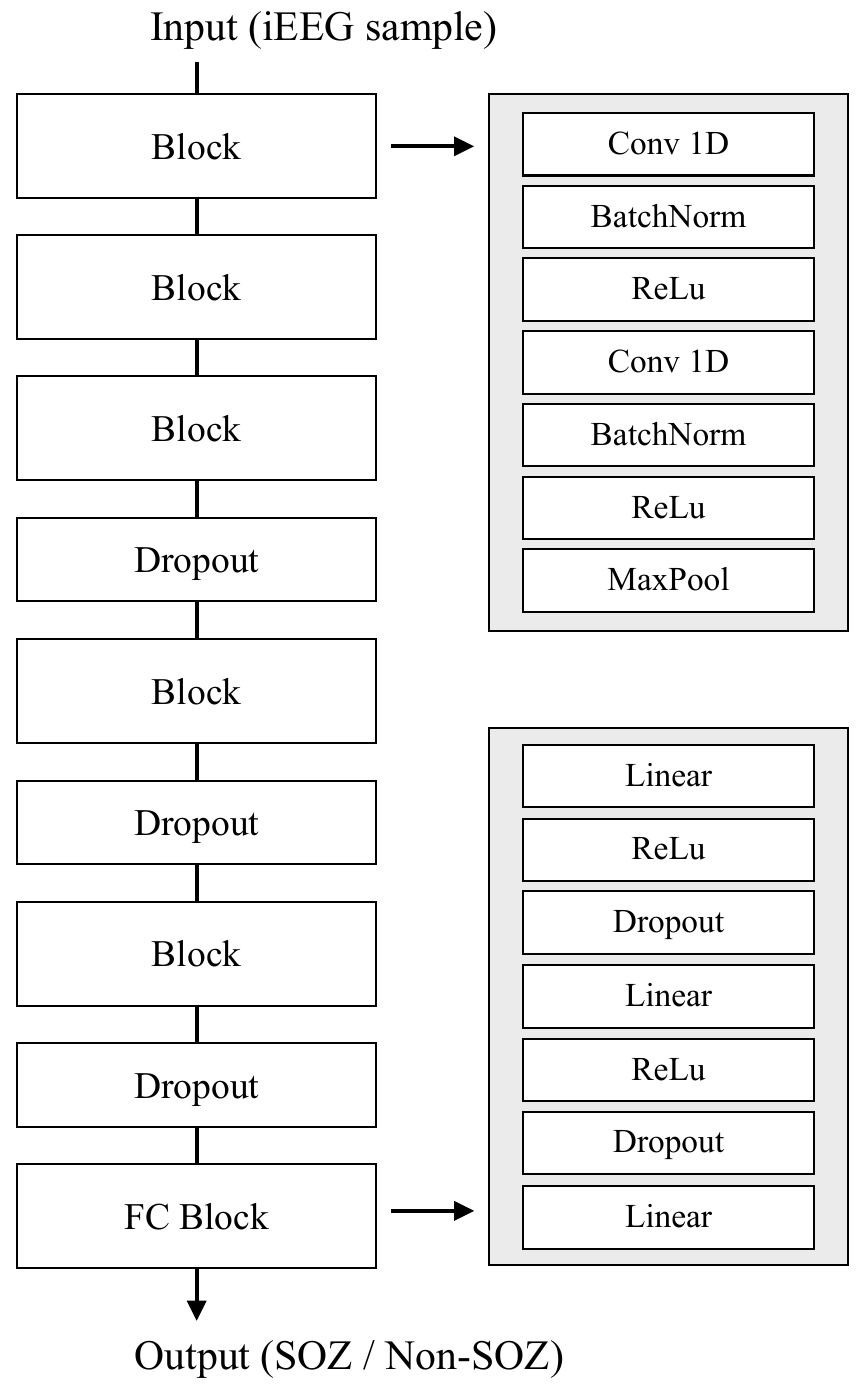}}
\caption{The model architecture of 1D-CNN.}
\label{fig_model}
\end{figure}

The architecture of the 1D-CNN model is shown in Figure~\ref{fig_model}. The model includes a total of ten one-dimensional convolutional layers, where the parameters of each layer are as follows (from top to bottom): 
Conv1d (in channels = 1, out channels = 32, kernel size = 3, stride = 1), Conv1d (32, 32, 3, 1), Conv1d (32, 64, 3, 1), Conv1d (64, 64, 3, 1), Conv1d (64, 128, 3, 1), Conv1d (128, 128, 3, 1), Conv1d (128, 256, 3, 1), Conv1d (256, 256, 3, 1), Conv1d (256, 256, 3, 1), Conv1d (256, 256, 3, 1). 
In the Block, the maxpool layer uses the MaxPool parameters (kernel size = 4, stride = 4) and the dropout layer uses the parameter of Dropout (0.3). 
The next part in the model is the fully convolutional neural network (FCNN), the parameters of each layer are as follows (from top to bottom): 
Linear(512, 256), Linear(256, 128), Linear(128, 64), Linear(64, 2). The dropout layers use the parameter of Dropout (0.5). 
The Cross Entropy loss function and Adam are used in the model training and fine-tuning phase, and the learning rate is tuned with ($5 \times 10^{-3}$, $1 \times 10^{-3}$, $5 \times 10^{-4}$, $1 \times 10^{-4}$, $5 \times 10^{-5}$, $1 \times 10^{-5}$).

%%%%%%%%%%%%%%%%%%%%%%%%
\begin{figure*}[htpb]
\centerline{\includegraphics[width=1.3\columnwidth]{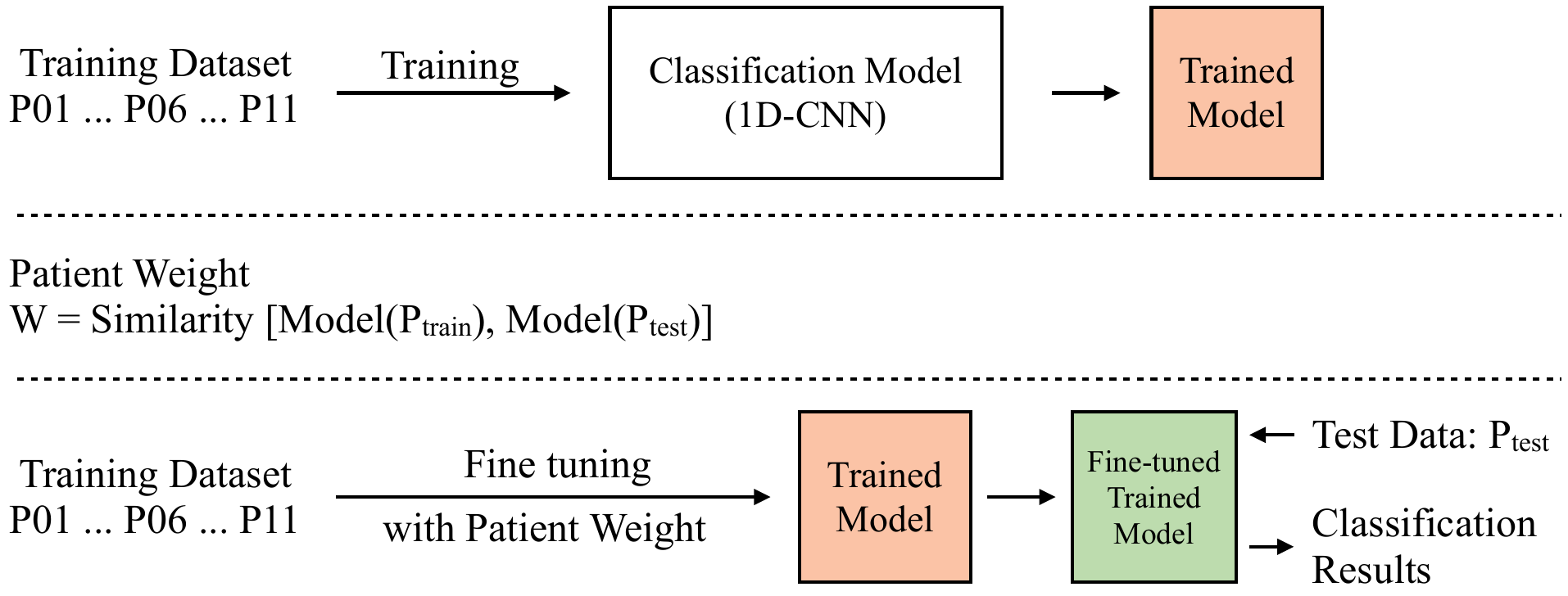}}
\caption{The overall structure of the patient weight method.}
\label{fig_method}
\end{figure*}
%%%%%%%%%%%%%%%%%%%%%%%%

\subsection{Patient weight method}
% In supervised learning, if there is a large difference in the distribution of test data and training data, the performance of the model will often be poor.
When applying supervised learning to medical data, the performance of the model in different test patients may vary greatly due to uncertain distribution differences between patient data.
In this work, we propose a patient weight method to solve the cross-patient problem in the iEEG data of epilepsy. The method first uses supervised learning to train the 1D-CNN model using data from multiple patients. Before making inferences on test patient data, different from standard supervised learning, we fine-tune a patient-specific model for each test patient.
During model fine-tuning, each training patient's data is assigned a weight based on its similarity to the test patient's data.
Use the training data with patient weight to fine-tune the pretrained model to make it more suitable for specific test patients and improve the performance of the model.

The specific description of the patient weight method is as follows:
The training data $D_{tr}$ contains data of $D_{P_{tr}^{i}} (i = 1, 2, \ldots, N)$ from multiple patients, in the experiment $N = 11$.
The test patient data of $D_{P_{te}}$ is used to evaluate the performance of the 1D-CNN model of $\mathbf{M}$ by the leave-one-patient-out method.
Next, to calculate the patient weights in the training data, we input the training data $D_{P_{tr}^{i}}$ and test data $D_{P_{te}}$ into the model $\mathbf{M}$ and save the output of the last convolutional layer of 
$O_{P_{tr}^{i}}  = \mathbf{M} ( D_{P_{tr}^{i}} )$
and 
$O_{P_{te}}  = \mathbf{M} ( D_{P_{te}} )$
%$O_{P_{tr}^{i}}$ and $O_{P_{te}}$ 
%
% The training data contains data from multiple patients
% The output of last convolutional layer in the model $\mathbf{M}$
% $O_{P_{tr}^{i}}  = \mathbf{M} ( D_{P_{tr}^{i}} )$
% $O_{P_{te}}  = \mathbf{M} ( D_{P_{te}} )$
Then, the weight $W_{P^{i}_{tr}}$ of each training patient data can be calculated as follows:
%$W_{P^{i}_{tr}} = Softmax ( \mathbf{MMD} (  O_{P_{tr}^{i}} , \  O_{P_{te}} )  )$.
%
\begin{equation}
\begin{aligned}
%W_{P^{i}_{tr}} = Softmax ( \mathbf{MMD} (  O_{P_{tr}^{i}} , \  O_{P_{te}} )  )  \nonumber
W_{P^{i}_{tr}} = \mathbf{MMD} (  O_{P_{tr}^{i}} , \  O_{P_{te}} ) \nonumber
\end{aligned}
\end{equation}
MMD is a kernel-based statistical test that is used to calculate the similarity between two data distributions.
In general, if the data for two patients are very similar, the weight of the training patient data is greater, and if the similarity is low, the weight of the training patient data is smaller.
%
% The patient weights $W_{P^{i}_{tr}}$ will be used in model fine-tuning so that data with similarity to the test patient will receive higher weights and vice versa.
Finally, the patient weights $W_{P^{i}_{tr}}$ will be used in the fine-tuning of the pre-trained model.
The samples of patient $P_{i}$ are multiplied by the weight of patient $W_{P^{i}_{tr}}$ when computing the loss function.
% so that patient data with similarity to the test patient receive higher weights, while the patient data with greater differences from the test patient receive lower weights.

\section{Experimental results}
The proposed patient weight method is evaluated using the Juntendo dataset, and in order to verify the cross-patient performance, leave-one-patient-out is used. Continuous iEEG data collected clinically were carefully labeled by clinical experts, and iEEG was used during the sleep stage for labeling, and it is ensured to be one hour after a seizure. Continuous iEEG data are divided into three-second samples, and each patient had two types of samples: SOZ and non-SOZ. Each patient contained 4,640 SOZ samples and 4,640 non-SOZ samples. In order to compare the effectiveness of the proposed method, standard supervised learning is first performed and then the weight method is verified under the same experimental conditions. The results of the two experiments will be explained separately.

\subsection{Results of standard supervised learning}
In standard supervised learning, using the leave-one-patient-out evaluation method, each training dataset contains data from ten patients.
% In model training, the batch size is 512 and the model epoch is 200 times.
The model is trained for 200 epochs with a batch size of 512.
The experimental results are shown in Table~\ref{tab_result}.
From the experimental results, it can be observed that the classification accuracy varies significantly across different test patients.
For example, for the patient with the worst classification result, as seen in the figure, the non-SOZ distribution in this patient is considerably different from that of other patients.
Such problems are widespread in medical data and difficult to improve.

%\begin{table*}[htbp]
%\caption{Summary of experimental results (accuracy).}
%\label{tab_result}
%\begin{center}
%\begin{tabular}{c | c | c | c | c | c | c | c | c | c | c | c | c}
%\toprule
%                 & P01    & P02   & P03    & P04   & P05    & P06    & P07   & P08    & P09    & P10    & P11   & Mean \\ \midrule
%Standard  & 83.45 & 48.76 & 91.82 & 44.35 & 74.11 & 76.68 & 62.20 & 76.58 & 92.34 & 50.77 & 26.92 & 66.18 \\ \midrule
%Multiscale & 87.99 & 59.93 & 93.89 & 69.70 & 81.90 & 82.95 & 78.45 & 81.11 & 97.88 & 64.62 & 51.06 & 77.23 \\ \midrule
%RBF          & 89.30 & 60.00 & 95.39 & 51.77 & 81.31 & 83.11 & 74.62 & 81.03 & 97.32 & 64.64 & 65.36 & 76.71 \\
%\bottomrule
%\end{tabular}
%\end{center}
%\end{table*}

\begin{table}[t]
\caption{Summary of experimental results with leave-one-patient-out evaluation (accuracy, \%).}
\label{tab_result}
\begin{center}
\begin{tabular}{c | c | c | c}
\toprule
       & Standard & Multiscale & RBF   \\ \midrule
P01 & 83.45      & 86.18         & 92.31 \\ \midrule
P02 & 48.76      & 66.67         & 51.94 \\ \midrule
P03 & 91.82      & 94.59         & 94.56 \\ \midrule
P04 & 44.35      & 45.41         & 45.84 \\ \midrule
P05 & 74.11      & 90.40         & 89.18 \\ \midrule
P06 & 76.68      & 83.71         & 82.86 \\ \midrule
P07 & 62.20      & 74.15         & 74.15 \\ \midrule
P08 & 76.58      & 80.63         & 80.51 \\ \midrule
P09 & 92.34      & 97.59         & 97.05 \\ \midrule
P10 & 50.77      & 66.67         & 71.82 \\ \midrule
P11 & 26.92      & 49.28         & 66.67 \\ \midrule
Mean &  66.18  & 75.93         & 76.99 \\
\bottomrule
\end{tabular}
\end{center}
\end{table}

\subsection{Results of patient weight method}

%\begin{figure*}[htpb]
%\caption{Random Channel Ordering Method.}
%\label{fig_result}
%\centerline{\includegraphics[width=2.0\columnwidth]{./fig/result-1.pdf}}
%\end{figure*}

\begin{figure}[t]
\centerline{\includegraphics[width=\columnwidth]{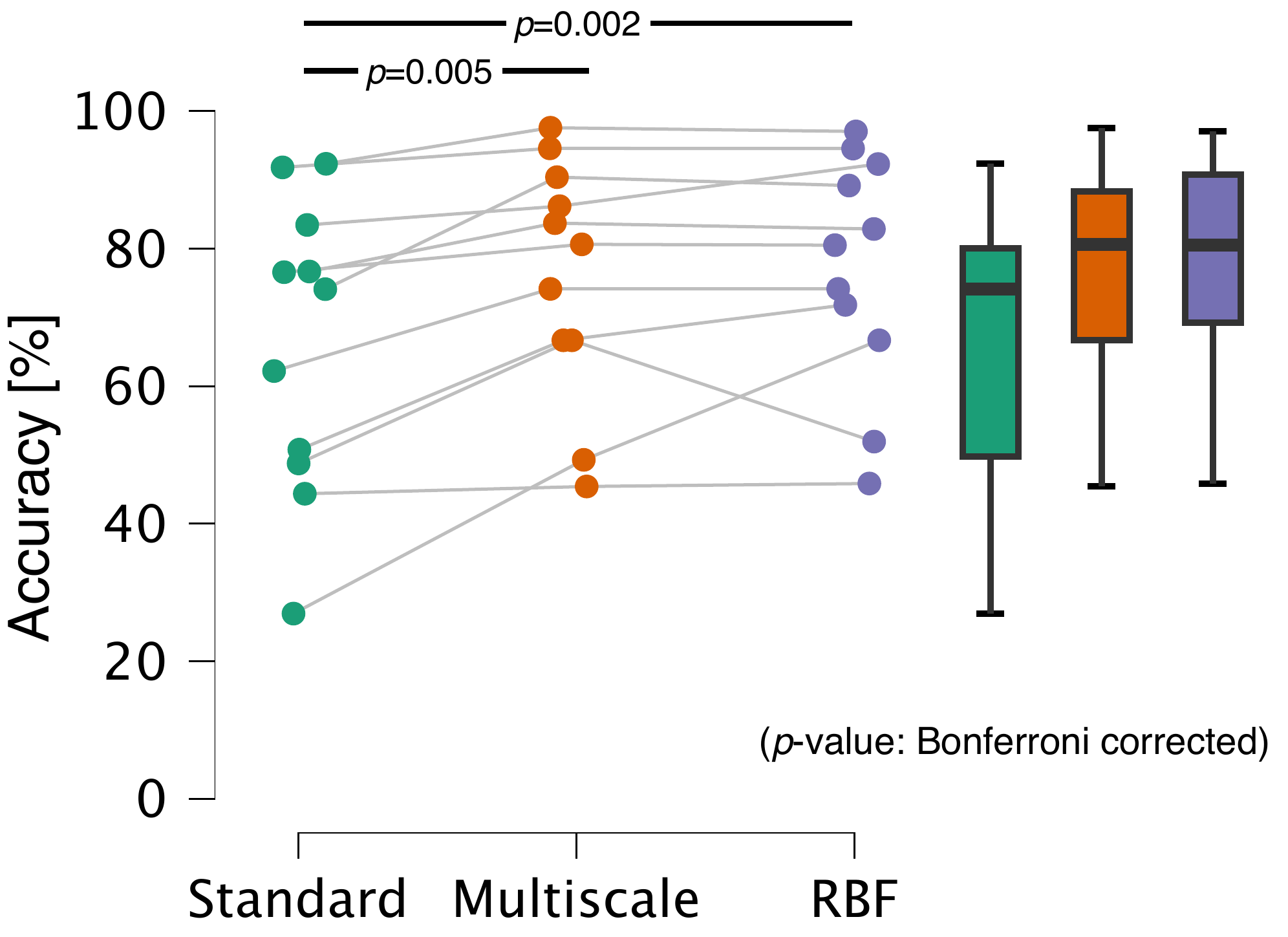}}
\caption{The experimental results with leave-one-patient-out evaluation (Accuracy \%).}
\label{fig_result}
\end{figure}

In this experiment, we first feed the training data and test data into the pretrained 1D-CNN model and extract the output from its final convolutional layer in the model.
The patient-specific weights are then calculated based on the extracted intermediate features.
Two variants of Maximum Mean Discrepancy (MMD)--Multiscale and RBF--are employed to calculate data similarity.
During the model fine-tuning phase, each sample is multiplied by the corresponding patient weight, so that samples with a distribution close to the test patients are given greater importance.
Since it is based on a trained model, the model fine-tuning is performed for only 5 epochs, and the batch size is 512, which is consistent with the previous experiment.
The experimental results are shown in Table~\ref{tab_result}, and the comparison of the experimental results between the standard supervised learning method and the patient weight method is shown in Figure~\ref{fig_result}.
Judging from the experimental results, the classification accuracy of all tested patients has improved to varying degrees, with an average improvement of more than eleven percent.
By fine-tuning a unique model for each test patient, training data similar to the test patient is given more weight, resulting in better model performance.

\section{Discussion}
This work aims to solve the cross-patient problem in medical data. Since medical data are composed of independent patients, each patient has different physical conditions and illnesses, resulting in certain differences in the distribution of the data of each patient.
Due to the existence of cross-patient problems, the performance of the model varies for different test patients in the model inference stage.
This depends on how similar the distribution of test patients is to the training data.
This article focuses on epilepsy. As can be seen in Figures~\ref{fig_visualization}, there are significant differences in the data distribution of different patients.
There are some methods~\cite{zhang2024cross, shafiezadeh2024systematic, zhang2024efficient, wang2022cross, matsubayashi2023identification} to solve the cross-patient problem in epilepsy data.
However, these methods usually require more complex calculations and more parameter adjustments, which makes them lack generality and are not conducive to clinical application.

In this work, we propose a method to improve the cross-patient problem in epilepsy data by adding patient weight.
The proposed method fine-tunes a specific model for each test patient without changing the model's architecture and training method.
The proposed method consists of two steps: first, a 1D-CNN model is trained based on the labeled iEEG data using the commonly used supervised learning method.
Next, we will use the patient weights and fine-tune the pretrained model from the first step.
Patient weights are calculated using the test patient data and each patient's data in the training data separately.
Instead of using the original iEEG data for calculation, the iEEG data is input into the pretrained model, the intermediate layer (last convolutional layer) output of the model is saved, and the MMD is used to calculate the similarity between the two patient data.
If the patient data in a certain training dataset is highly similar to the test patient data, the weight of these patient data is also larger; otherwise, the weight is smaller.
Finally, the training dataset is used together with the patient weights to fine-tune the pretrained model, thereby providing a specific model for each test patient.

In the experiments, an epilepsy iEEG dataset is used to evaluate the performance of the proposed method.
In standard supervised learning, leave-one-patient-out is used to evaluate the general performance of the 1D-CNN model.
From the experimental results shown in the table~\ref{tab_result}, 
we can see that there are huge performance differences between the models in different patient data.
%The experimental results are shown in the table~\ref{tab_result}.
This result is also consistent with the distribution differences of different patient data shown in Figure~\ref{fig_visualization}.
%
%From the experimental results, the performance of each patient has been improved under the two methods of calculating similarity (MMD \& Multiscale, RBF), and the average increase of more than 11 \% in accuracy.
In the patient-weighted method, the patient weight is calculated by MMD (Multiscale \& RBF).
% The weights are calculated using Algorithm 1.
The generalizability of the 1D-CNN model in different patients is improved by using fine-tuning and patient weight, with an average improvement of more than 11\% in accuracy.

%%%%%%%%%%%%%%%%%%%%%%%%%%%%%%%%%%%%
% In this work, we propose a method based on patient weight fine-tuning to improve the cross-patient problem existing in the epilepsy data. The method consists of two steps, the first step is to train the model using a training dataset by the common supervised learning, and a one-dimensional convo- lutional neural model (1D-CNN) is used as the classifier for SOZ and Non-SOZ classification task. In the second step, we fine-tune the trained model in the first step using the training data, and each patient in the training data is assigned a patient weight. The patient weights are calculated using the output of the model’s intermediate layers (last convolutional layer) between test patient data and each training patient data. In the experiment, we calculated the similarity between the test patient and each training patient by using the maximum mean discrepancy (MMD) [24] method. The similarity is then processed through softmax and used as the patient weight for model fine-tuning. In general, if the similarity between the test patient and one of the training patient data is high, the weight of this patient data will increase, otherwise, it will decrease. In the experiment, the leave-one-patient-out is used to evaluate the performance of the proposed method. From the experimental results, the performance of each patient has been improved under the two methods of calculating similarity (MMD \& Multiscale, RBF), and the average increase of more than 11 \% in accuracy.
%%%%%%%%%%%%%%%%%%%%%%%%%%%%%%%%%%%%

\section{Conclusion}
Machine learning plays an increasingly important role in assisting medical diagnosis, but the cross-patient problem that is prevalent in medical data causes machine learning to exhibit different performance among different patients.
% In response to the widespread cross-patient problem in medical data, 
In this work, we propose a patient weight method to improve the generalization performance of machine learning.
% we propose a patient weight method.
For each test patient, before using the model to make inferences, we first calculate the weight between the test patient and the training patient and use this patient's weight to fine-tune the trained model. 
In this way, patient data in the training with a similar distribution to the test patient data will be valued in model fine-tuning, while those with a different distribution will be disregarded.
From the experimental results, the fine-tuned model has shown varying degrees of improvement in the classification of all patient data.

\section*{Acknowledgment}
This work was supported by JST CREST Grant Number JP-MJCR1784.

% \begin{thebibliography}{1}

% \bibitem{1}
% G.~Eason, B.~Noble, and I.~N.~Sneddon, ``On certain integrals of
% Lipschitz-Hankel type involving products of Bessel functions,''
% \emph{Phil. Trans. Roy. Soc. London,} vol. A247, pp. 529-551, April
% 1955.

% \bibitem{2}
% J.~Clerk~Maxwell, \emph{A Treatise on Electricity and Magnetism,}
% 3$^{\rm rd}$ ed., vol. 2. Oxford: Clarendon, 1892, pp.68-73.

% \bibitem{3}
% I.~S.~Jacobs and C.~P.~Bean, ``Fine particles, thin films and exchange
% anisotropy,'' in \emph{Magnetism,} vol. III, G.T. Rado and H. Suhl,
% Eds. New York: Academic, 1963, pp. 271-350.

% \bibitem{4}
% K.~Elissa, ``Title of paper if known,'' unpublished.

% \bibitem{5}
% R.~Nicole, ``Title of paper with only first word capitalized,''
% \emph{J. Name Stand. Abbrev.,} in press.

% \bibitem{6}
% Y.~Yorozu, M.~Hirano, K.~Oka, and Y.~Tagawa, ``Electron spectroscopy
% studies on magneto-optical media and plastic substrate interface,''
% \emph{APSIPA Transl. J. Magn. Japan,} vol. 2, pp. 740-741, August 1987
% [\emph{Digests 9$^{\rm th}$ Annual Conf. Magnetics Japan,} p. 301,
% 1982].

% \bibitem{7}
% M.~Young, \emph{The Technical Writer's Handbook.} Mill Valley, CA:
% University Science, 1989.

% \end{thebibliography}

%\printbibliography

%\clearpage

\bibliographystyle{IEEEtran}
\bibliography{paper.bib}

%\usepackage[style=plain]{biblatex}  % 你也可以选择 apa, ieee, etc.
%\addbibresource{PaperSample_Guideline_tex_modified.bib}

\end{document}